\documentclass[10pt]{article}
\usepackage{cite}
\usepackage{graphicx}
\usepackage{amsmath,amssymb} 
\usepackage{color}
\usepackage{multirow}
\usepackage{hyperref}
\usepackage{geometry} 
\title{AIRNet: Self-Supervised Affine Registration \\ for 3D Medical Images using Neural Networks} 
\author{Evelyn Chee \\ \href{mailto:evelyn.chee@biomind.ai}{evelyn.chee@biomind.ai} 
   \and Zhenzhou Wu \\ \href{mailto:joe.wu@biomind.ai}{joe.wu@biomind.ai}}
\date{} 

\begin{document}

\maketitle

\begin{abstract}

In this work, we propose a self-supervised learning method for affine image 
registration on 3D medical images. Unlike optimisation-based methods, our affine 
image registration network (AIRNet) is designed to directly estimate the 
transformation parameters between two input images without using any metric, 
which represents the quality of the registration, as the optimising function. But
since it is costly to manually identify the transformation parameters between
any two images, we leverage the abundance of cheap unlabelled data to generate
a synthetic dataset for the training of the model. Additionally, the structure of 
AIRNet enables us to learn the discriminative features of the images which are 
useful for registration purpose. Our proposed method was evaluated on magnetic 
resonance images of the axial view of human brain and compared with the 
performance of a conventional image registration method. Experiments 
demonstrate that our approach achieves better overall performance on registration 
of images from different patients and modalities with 100x speed-up in execution 
time.

\end{abstract}

\section{Introduction}
\label{sec:introduction}

Image registration \cite{imageregistration1} is the process of aligning different 
images into the same coordinate frame, enabling comparison or integration of
images taken at different times, from different viewpoints, or by different sensors. 
It is used in many medical image analysis tasks. Several medical image registration 
studies have been conducted \cite{imageregistration2,imageregistration3,imageregistration4,imageregistration5}
and toolkits such as SimpleITK \cite{simpleitk,simpleitk2}, ANTs \cite{ants} and 3D 
Slicer \cite{3dslicer} have been developed. Typically, registration processes 
implemented in those tools are performed by iteratively updating transformation 
parameters until a predefined metric, which measures the similarity of two images 
to be registered, is optimised. These conventional methods have been achieving 
decent performance, but their applications are limited by the slow registration 
speed. This is mainly because the iterative algorithm is optimising the cost function 
from scratch for every new registration task instead of utilising the information 
obtained from previous experiences.  

To overcome this issue, many recent works done on medical image registration 
have proposed methods based on deep learning approaches, motivated by the 
successful applications of convolutional neural network (CNN) in the computer vision 
field. Wu et al. \cite{featurerepresentation} use a convolutional stacked auto-encoder 
to learn the highly discriminative features of the images to be registered. However,
the extracted features might not be optimal for registration purpose as the extraction 
is done separately for each image pairs. Also, the method proposed is not an 
end-to-end deep learning strategy as it still relies on other feature-based registration 
methods to find the mapping between the two images. 

Subsequently, several works on end-to-end unsupervised registration using CNN 
have been presented. Similar to the conventional image registration methods, 
the strategy proposed by de Vos et al. \cite{similaritymetric} and Shan et al. 
\cite{photometricdiff} does not require any ground truth label on the image pairs 
to be registered. A predefined metric is used as the loss function and back-propagated 
to the CNNs to learn the optimal parameters of the network that minimises the error. 
This strategy is implementable by using the spatial transformer network introduced 
by Jaderberg et al. \cite{stn}, which enables neural networks to spatially transform 
feature maps and is fully differentiable. To ensure the satisfactory performance of 
these frameworks, a good optimising metric must be defined. This could be a 
potential drawback as different metrics have their pros and cons and the suitability 
of a metric varies from task to task.  

On the other hand, Miao et al. \cite{transformation} use CNN regression to directly 
estimate the transformation parameters and there is no selection of the right
optimising metric involved. The models were then trained on synthetic X-ray
images as they provide ground truth labels without the need of manual annotation.  
Although higher registration success rates than conventional methods have been
achieved, in their framework, six regressors were trained and applied in a 
hierarchical manner instead of estimating all the parameters simultaneously.

\begin{figure}
    \centering
    \includegraphics[width=0.9\textwidth]{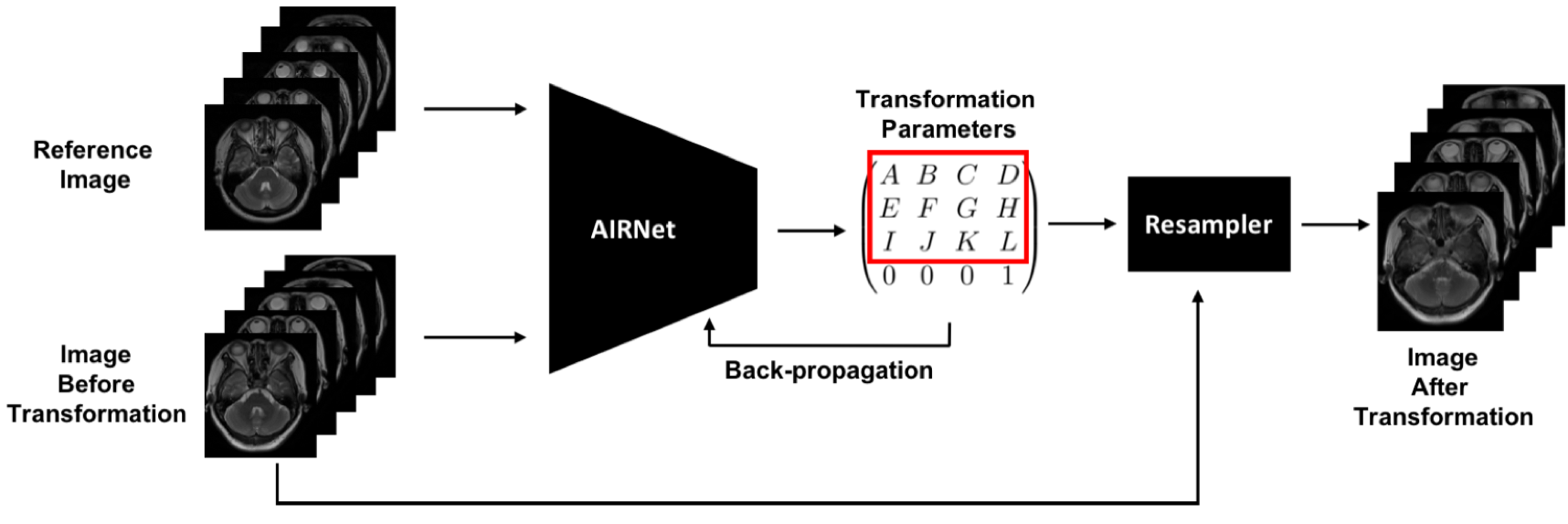}
    \caption{The workflow of our image registration framework. The AIRNet takes two 
    images and outputs the transformation parameters ({\it red box}). The image to 
    be transformed is then warped by the resampler using the transformation matrix. 
    During training, the loss function, which is the mean-squared error of the estimated 
    transformation parameters, will be back-propagated to the network}
    \label{fig:framework}
\end{figure}

In this paper, we build a system (Fig.~\ref{fig:framework}) for medical image 
registration, which also utilises CNN to calculate the transformation parameters 
between two 3D images. Unlike the unsupervised learning methods that
maps the images to a scalar-valued metric function, our affine image registration 
network (AIRNet) requires two 3D images to be registered (input values) and the 
transformation parameters (target values). Here, we use a twelve-dimensional vector 
capturing 3D affine transformation as the label of each input data. But since it is 
impractical to manually identify the transformation parameters between any two 
images, we explore a self-supervised learning method and leverage the abundance 
of cheap unlabelled data to generate a synthetic dataset for the training of the model. 
Through direct estimation of transformation matrix, our method performs one-shot 
registration instead of hierarchical manner as presented by Miao et al \cite{transformation}. 

Furthermore, the structure of the AIRNet enables us to learn the discriminative 
features of the images which are useful for registration purpose. Additionally, the 
features allow us to obtain generalised representations of different image modalities. 
For instance, different set of a brain scan can be generated based on different settings 
of magnetic resonance (MR) brain imaging protocol, rendering the intensity of the 
brain tissue to differ across modalities. But for brain volumes which are aligned, the 
representation of these images would be similar as the registration focuses on 
geometric information. 

In short, we develop a deep learning framework for 3D image registration which 
achieves better overall performance at 100x faster speed in execution as compared 
to some conventional methods. Our framework is applicable for two main areas 
in medical imaging: 1) registration of images from different patients to identify
different anatomical regions of the body, and 2) registration of images from 
different modalities to integrate useful information across different type of data. 
The rest of the paper is organised as follows: Sect.~\ref{sec:problem} defines 
our problem; Sect.~\ref{sec:methods} introduces our proposed method; 
Sect.~\ref{sec:experiments} presents the experimental results. 

\section{Problem Statement}
\label{sec:problem}

In this paper, the image registration system focuses on 3D affine transformation, 
which includes translation, scaling, rotation and shearing. Each of these 
transformation can be parameterised by a matrix \(T\) as shown in 
Fig.~\ref{fig:framework}. Using these matrices, we map an input coordinate to 
another coordinate and obtain the desired transformation effect. In other words, 
for any pixel coordinates \(\vec{x}\) in a 3D image, the new pixel location can be 
computed by \(\vec{x}' = T\vec{x}\). The values of the unknowns in the matrix 
indicate the amount of translation, scaling, rotation and shearing. Replacing these 
letters by different values would lead to various compositions of these 
transformations. Hence, it is sufficient to represent the relationship between two 
3D images using a twelve-dimensional vector \(\vec{t}\), which is the flatten copy 
of the top three rows of matrix \(T\). 

As shown in Fig.~\ref{fig:framework}, the inputs of our framework are the two 
images to be registered. We denote the reference image as \(I_{ref}\) and the
other image to be shifted to match the reference as \(I_{\vec{t}}\), where \(\vec{t}\) 
is the unknown ground truth transformation parameters. The registration problem 
is then formulated as 
\begin{equation}
\vec{t} = \text{AIRNet}(I_{ref}, I_{\vec{t}}),
\end{equation}
where the AIRNet is trained to find the mapping from the input images to the 
transformation parameters. The design of the network and methods used to find 
the optimal weights in the AIRNet will be discussed in the next section.

\section{Methods}
\label{sec:methods}

This section describes the architecture of the AIRNet and the training procedure.

\begin{figure}
    \centering
    \includegraphics[width=0.9\textwidth]{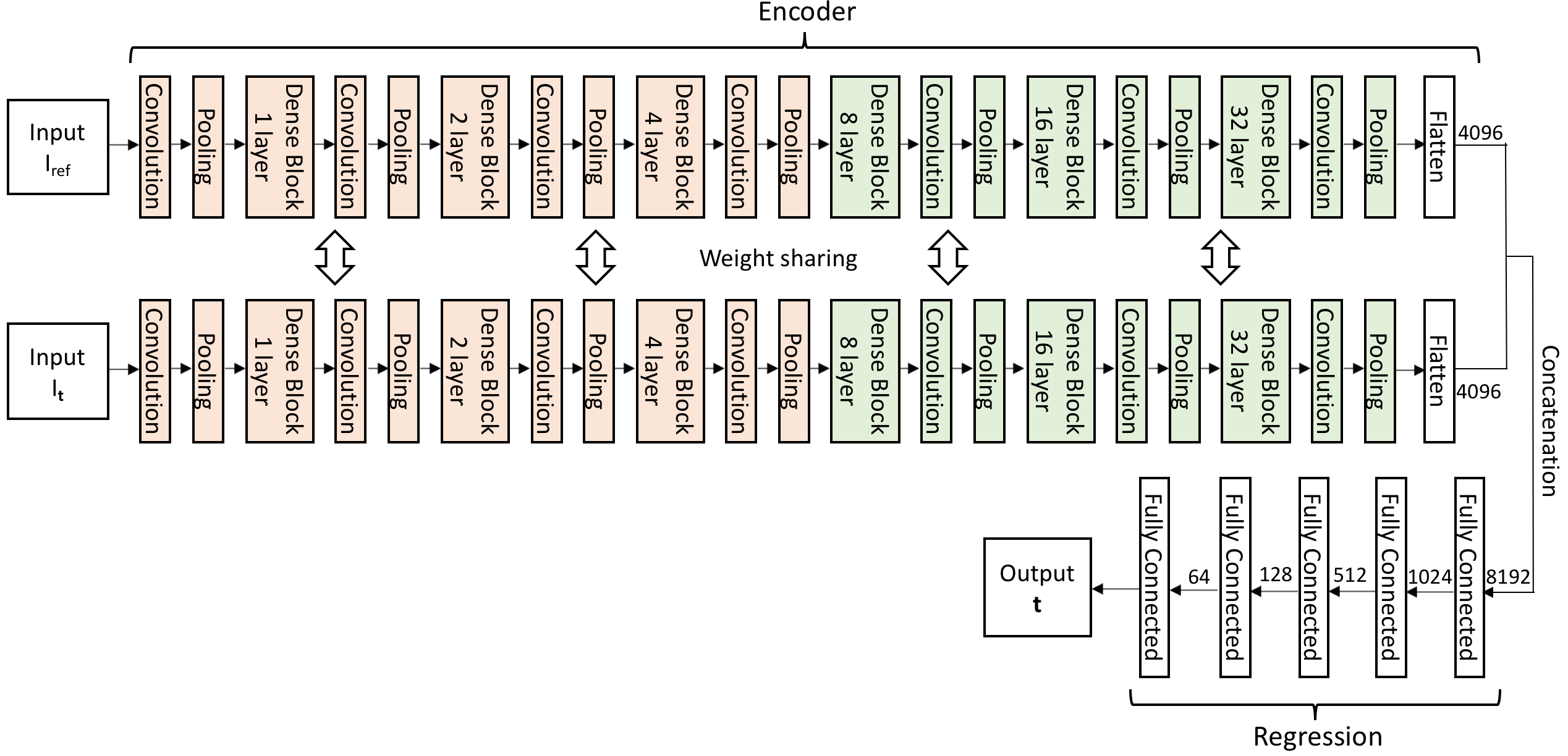}
    \caption{Architecture of AIRNet. Two separate paths are used for $I_{ref}$ and 
    $I_{\vec{t}}$. The layers before concatenation act as an encoder to learn 
    representations of the data and each block from the two pathways share the 
    same parameters. The first part of the encoder ({\it orange blocks}) consists of 
    2D filters whereas 3D filters are used in the remaining part ({\it green blocks}). 
    The outputs of the encoder are flatten and concatenated before passed into the 
    fully connected layers}
    \label{fig:network}
\end{figure}

\subsection{Network Architecture} 

The AIRNet is designed to model the affine transformation relationship between 
two input images as shown in Fig~\ref{fig:network}. There are two main components 
in the network: the encoder part which captures discriminative features of the input 
images and the regression part which finds the non-linear relationship between the 
features and the corresponding transformation parameters.

For the encoder, it learns a new representation of the input data optimally for 
registration purpose, and these representations allow us to gain insights to the 
image dataset. From Fig.~\ref{fig:network}, the encoder is composed of two 
pathways, one for each input image. The corresponding blocks from both pathways 
share the same parameters. Weight sharing between blocks allows us to learn
the same encoder for feature extraction of both images, as well as reduces the
number of parameters in the first part of the network by half to minimise the risk
of overfitting.

The architecture of the encoder is adapted from DenseNet \cite{densenet}, in which
the building blocks have skip connections to allow integration of high and low-level 
feature information. An additional property of our network is the mixture of 2D 
and 3D filters which aims to resolve the anisotropicity of the 3D medical images, 
i.e. the depth of the input image is much smaller compared to its height and width. 
So, in the initial part of the encoder, 2D filters are used in the convolutional and 
pooling layers. The height and width of the feature maps are shrunk as the
effect of the 2D pooling layers. Consequently, the size of the features in the 
height and width directions will eventually be similar to that in the depth 
direction, after which 3D convolutional and pooling layers will be used.

Each input image firstly undergoes a $3 \times 3$ convolutional layer 
and is downsized using a $2 \times 2$ max-pooling layer. The dense blocks 
are made of multiple 2D (resp., 3D) convolutional layers with a growth rate of 
8 with each layer taking the input of the previous feature map as input. The 
number of layers in each dense block are indicated in Fig.~\ref{fig:network}. 
More layers are added as the network goes deeper because deeper layers 
generally correspond to more specific features and hence, require more channels 
to capture. Next, the transition layers, which are the layers between dense
blocks, consist of an $1 \times 1$ (resp., $1 \times 1 \times 1$) convolutional layer, 
a batch normalisation layer and a Rectified Linear Unit (ReLU) function followed 
by a $2 \times 2$ (resp., $2 \times 2 \times 2$) max-pooling layer with a 
stride of 2. 

The outputs of both flatting layers are then concatenated and given as input to 
the regression part. The fully connected layers are where the high-level reasoning
is done and the non-linear relationship between features extracted from the 
encoder is decided. These layers consist of linear layers, batch normalisation 
layers and ReLU functions. The regression model will then return the transformation 
parameters that align the input images. 

\subsection{Training}

Our proposed framework is experimented on with MR images of the axial view of 
human brain from our hospital database and we trained the network in a supervised 
manner by minimising a loss between the actual and predicted transformation 
parameters. However, information regarding the transformation matrix between 
any two of the images are not available. Thus, we used a self-supervised learning
approach and trained the network on synthetic MR images. The process of generating 
this training dataset should provide reliable ground truth labels with little manual 
annotation. So, for every brain scan, a random transformation matrix $T$ was 
generated. The range of each transformation type could be set according to the aim of 
each particular experiment. In our case, the constraints of the parameters are as follows 
($x$, $y$ and $z$-axis corresponds to the width, height and depth direction): angle of 
rotation about $z$-axis ranges from -0.8 to 0.8 radians; translation ranges from -0.15 
to 0.15 along $x$ and $y$-axis and -0.2 to 0.2 along $z$-axis; scaling across all axes 
ranges from 0.8 to 1.3. The matrix $T$ was then applied onto the image to produce an 
affine transformed version of the brain scan. These two images are now the two input 
images of the network, $I_{\vec{t}}$ and $I_{ref}$ respectively where the ground truth 
label of the input pair, $\vec{t}$, is the flatten copy of the top three rows of matrix $T$.  

There are some pre-processing methods implemented on the images. Firstly, 
to ensure performance, the range of pixel values in different input images is unified
by normalising the pixel values to a range between zero and one. Secondly, the 
images were resized and resampled so that all images have the same dimension, 
which in this case is $20 \times 320 \times 320$ pixels.

\begin{figure}
    \centering
    \includegraphics[width=0.45\textwidth]{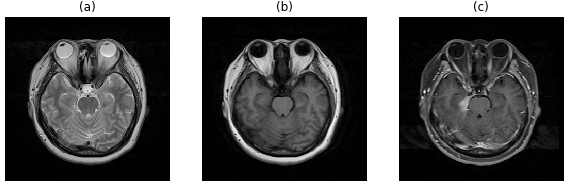} 
    \caption{MR brain images in the axial plane which are: (a) T2-weighted, 
    (b) T1-weighted, (c) contrast-enhanced T1-weighted}
    \label{fig:modalities}
\end{figure}

As introduced earlier, the encoder in the AIRNet is able to extract a generalised 
representation of different image modalities. Three modalities used in this study 
are T2-weighted, T1-weighted and contrast-enhanced T1-weighted 
(Fig.~\ref{fig:modalities}). Although pixel intensities of the same brain tissue can 
vary in different modalities, the encoder can capture general features across 
modalities. The selection of most discriminative features, which concisely 
describe complex morphological patterns of the brain, addresses the concern of 
the registration method not being able to scale well to various image modalities. 
To test the effectiveness of our encoder, three training datasets were prepared:
\begin{itemize}
    \item [$\bullet$] $S_{1,2,c}$: All three modalities are included in the dataset and 
    each image pair, $I_{ref}$ and $I_{\vec{t}}$ are of the same modality.
    \item [$\bullet$] $S_{1,2}$: Each image pair are of the same modality but only 
    images of T2 and T1-weighted are used.
    \item [$\bullet$] $D_{1,2}$: For each patient, images of T2 and T1-weighted are 
    already aligned because both are acquired through one scan and hence, we 
    constructed a dataset which have these two different modalities within each 
    image pair.
\end{itemize}

Using these datasets, we trained three models using Tensorflow \cite{tensorflow}.
The objective function to be minimised during the training is defined as:
\begin{equation}
Loss = \frac{1}{K}\sum_{i=1}^{K}||\vec{t}^{[i]} - \text{AIRNet}(I_{ref^{[i]}}, I_{\vec{t}^{[i]}}; 
\vec{W})||_2^2,
\end{equation}
where $K$ is the number of training sample, $\vec{W}$ is a vector of weights in 
the network that needs to be learned, $I_{ref^{[i]}}$ and $I_{\vec{t}^{[i]}}$ are the 
$i$-th training input pairs, $\vec{t}^{[i]}$ is the label of the $i$-th training sample, 
AIRNet$(I_{ref^{[i]}}, I_{\vec{t}^{[i]}}; \vec{W})$ is the output of the network 
parameterised by $\vec{W}$ on the $i$-th training sample. The weights $\vec{W}$ 
are learned using the Adam optimiser algorithm \cite{adamoptimizer}.

\section{Experiments and Results}
\label{sec:experiments}

As described in the previous section, the AIRNet estimates the transformation
matrix given two images. This matrix is then feed into a resampler which wraps
one of the image to the other. Later, registration performance of the trained neural 
network will be assessed and compared with a conventional registration method.

\subsection{Conventional Registration Algorithm}

The conventional registration method we used is the affine transformation function 
available under the SimpleITK \cite{simpleitk,simpleitk2} registration framework. 
Mattes mutual information \cite{mattes} with the default settings is selected as the 
similarity metric and linear interpolator is used. For the optimiser, we adopted 
regular step gradient descent with learning rate of 0.005, minimum step of 
$1 \times 10^{-10}$ and number of iterations as 10000.

\subsection{Evaluation}
 
We evaluate the trained models on a MR brain image dataset with ground truth 
segmentation masks which were prepared from our hospital database. This 
dataset consists of twelve subjects and each has T2 and T1-weighted MR images. 
For each subject, manual anatomical segmentation of the brain has been done. 
Since the image of both modalities are aligned, the segmentation masks are 
applicable to both images. 

For any two images which are perfectly registered, the brain parts represented by  
matching pixels should be consistent in both images. Hence, we could use the 
similarity between the segmentation masks of the reference and registered 
images to evaluate the performance of the registration methods. The following 
metrics are used throughout all experiments to compare two masks, $A$ and $B$:
\begin{itemize}
    \item [$\bullet$] Jaccard index ($Jac$): It is used for measuring the similarity 
    of the masks and is defined as the size of their intersection divided by the size of 
    their union:
    \begin{equation}
    Jac(A,B) = \frac{|A \cap B|}{|A \cup B|}.
    \end{equation}
    \item [$\bullet$] Modified Hausdorff distance \cite{hausdorff} ($d_H$): It 
    measures the distance between two masks and is defined as the average of all 
    the distances from a boundary point in one mask to the closest boundary point in 
    the other mask:
    \begin{equation}
    d_H(A,B) = \max\{\frac{1}{|P_A|}\sum_{a \in P_A}\inf_{b \in P_B} d(a,b), 
    \frac{1}{|P_B|}\sum_{b \in P_B}\inf_{a \in P_A} d(a,b)\},
    \end{equation}
    where $P_A$ and $P_B$ are the set of boundary points in $A$ and $B$ 
    respectively, $d$ represents the Euclidean distance, and $|\cdot|$ measures 
    the number of elements in the set.
\end{itemize}
The third metric reported is the running time, which records the total time required, 
in seconds, to calculate the transformation parameters of the two images. The results 
indicated are based on the implementation of the algorithms on a Intel Xeon CPU 
E5-2630 v4 2.2GHz.
     
\subsection{Registration Results}

\begin{figure}
    \centering
    \includegraphics[width=0.9\textwidth]{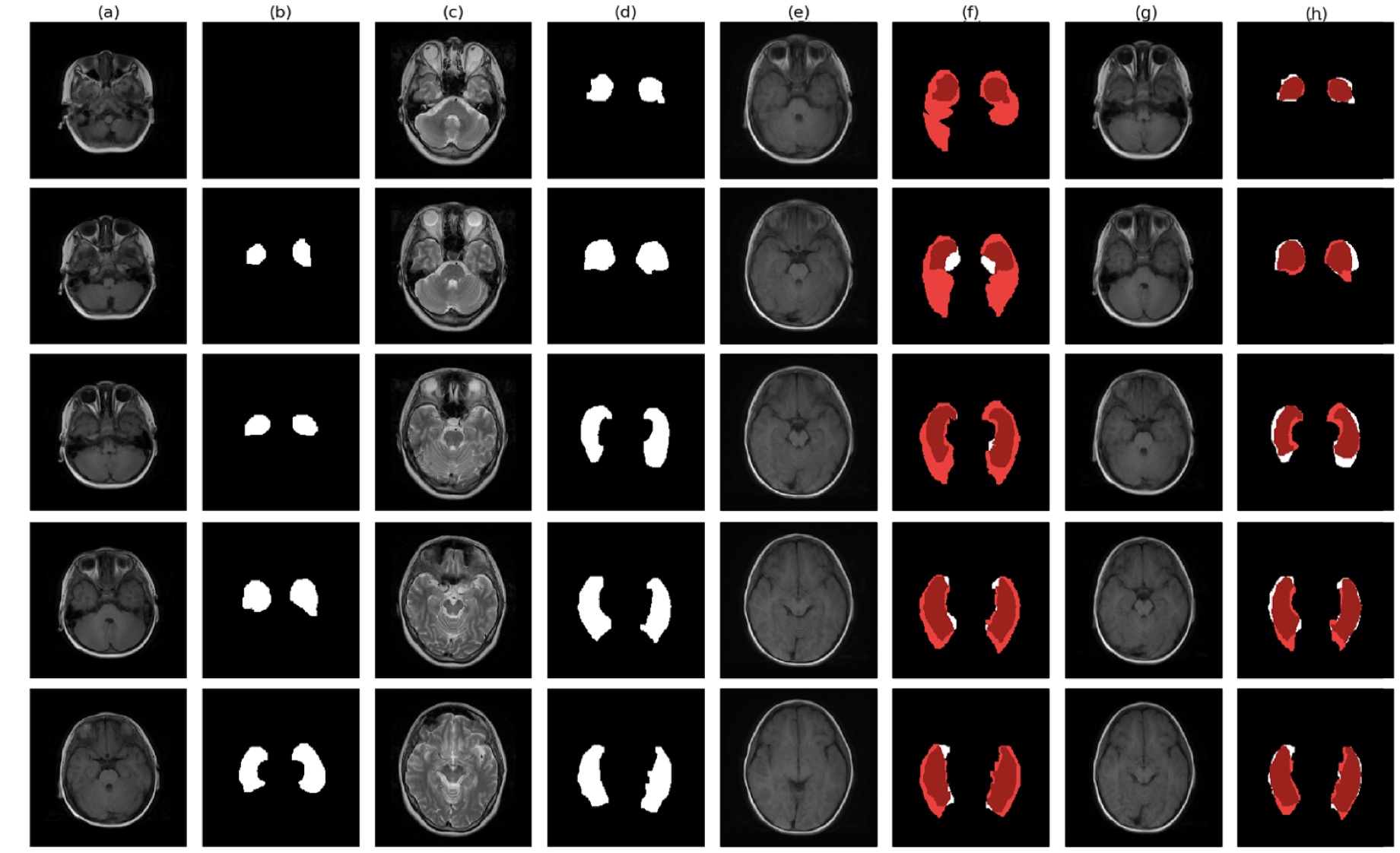}
    \caption{Illustration of the brain registration performance of AIRNet-$S_{1,2,c}$ 
    and the conventional registration algorithm. The snapshots shown are slice 5 to 9 
    of the following: (a) Image to be registered $I_{\vec{t}}$; (b) Ground truth segmentation 
    mask of $I_{\vec{t}}$; (c) Reference image $I_{ref}$; (d) Ground truth segmentation 
    mask of $I_{ref}$. (e) and (g) are $I_{\vec{t}}$ warped by SimpleITK and
    AIRNet respectively. The red masks in (f) and (h) are the warped ground truth 
    segmentation mask which corresponds to (e) and (g) while the white masks are 
    the ground truth segmentation mask of $I_{ref}$}
    \label{fig:comparison}
\end{figure}

We firstly select one subject $P$ from the test dataset to act as a reference point. 
In other words, the MR images of this subject will be used as $I_{ref}$ in the 
following two experiments: 
\begin{itemize}
    \item [$\bullet$] $E_p$ (across patients registration): Registering T1-weighted  
    image of the remaining subjects to T1-weighted image of subject $P$;
    \item [$\bullet$] $E_{p,m}$ (across patients and modalities registration): Registering 
    T1-weighted image of the remaining subjects to T2-weighted image of subject $P$.
\end{itemize}
The estimated transformation parameters are then fed into the resampler to wrap 
the corresponding segmentation masks. We firstly evaluate the metrics on the
segmentation mask of one particular brain part, which is the temporal lobe and the
performance of our neural registration models and the conventional method are 
shown in Table~\ref{table:one_reg}. Our proposed models have relatively constant
execution speed and are 100x faster than the conventional method. At the same 
time, they still achieve better overall registration performance for this region. 
Fig.~\ref{fig:comparison} illustrates an example of the registration results from 
experiment $E_{p,m}$. 

\setlength{\tabcolsep}{4pt}
\begin{table}
    \begin{center}
    \caption{Performance of the conventional methods and our proposed framework 
    when evaluated on the temporal lobe. The metrics are 1) mean Jaccard index 
    across all subjects, 2) mean modified Hausdorff distance across all subjects, and 
    3) average and standard deviation of running time per registration. For the two
    respective experiments, the rows list results before registration, with registration 
    using SimpleITK, and results obtained using our three trained models. The best 
    results are highlighted in bold.}
    \label{table:one_reg}
    \begin{tabular}{llccc}
    \hline\hline
    \noalign{\smallskip}
    Experiment & Method & \quad\quad $Jac$ \quad\quad & \quad\quad $d_H$ \quad\quad & Running Time \\
    \noalign{\smallskip}
    \hline\hline
    \noalign{\smallskip}
    - & No registration & 0.491 & 2.270 & - \\
    \noalign{\smallskip}
    \hline
    \noalign{\smallskip}
    \multirow{5}{*}{$E_p$} & SimpleITK \cite{simpleitk,simpleitk2} & 0.561 & 1.876 & 166.537$\pm$89.844 \\
    & AIRNet-$S_{1,2,c}$ & \bf{0.613} & \bf{1.431} & 0.785$\pm$0.027 \\
    & AIRNet-$S_{1,2}$ & 0.586 & 1.546 & 0.778$\pm$0.036 \\
    & AIRNet-$D_{1,2}$ & 0.599 & 1.512 & 0738$\pm$0.016 \\
    \noalign{\smallskip}
    \hline
    \noalign{\smallskip}
    \multirow{5}{*}{$E_{p,m}$} & SimpleITK \cite{simpleitk,simpleitk2} & 0.524 & 1.978 & 216.848$\pm$114.369 \\
    & AIRNet-$S_{1,2,c}$ & \bf{0.615} & \bf{1.459} & 0.793$\pm$0.034 \\
    & AIRNet-$S_{1,2}$ & 0.577 & 1.574 & 0.772$\pm$0.028 \\
    & AIRNet-$D_{1,2}$ & 0.601 & 1.475 & 0.755$\pm$0.022 \\
    \noalign{\smallskip}
    \hline \hline
    \end{tabular}
    \end{center}
\end{table}
\setlength{\tabcolsep}{1.4pt}

\begin{figure}
    \centering
    \includegraphics[width=\textwidth]{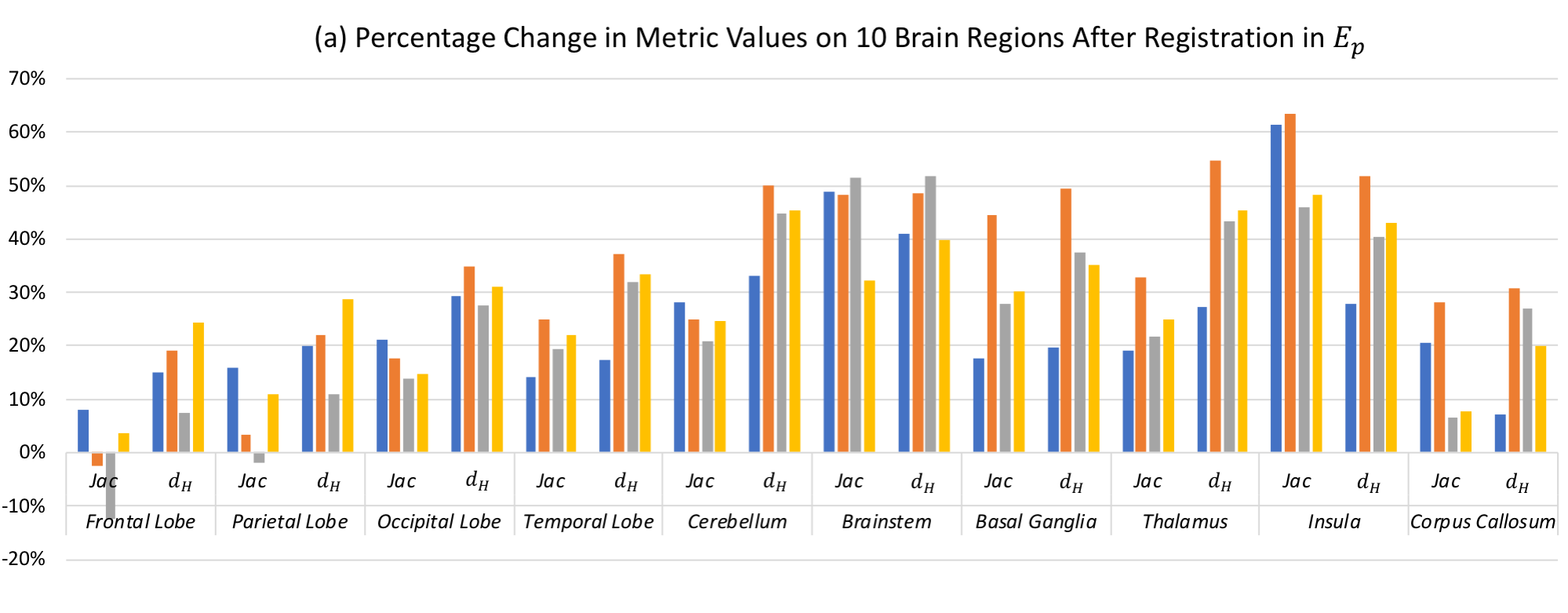} 
    \includegraphics[width=\textwidth]{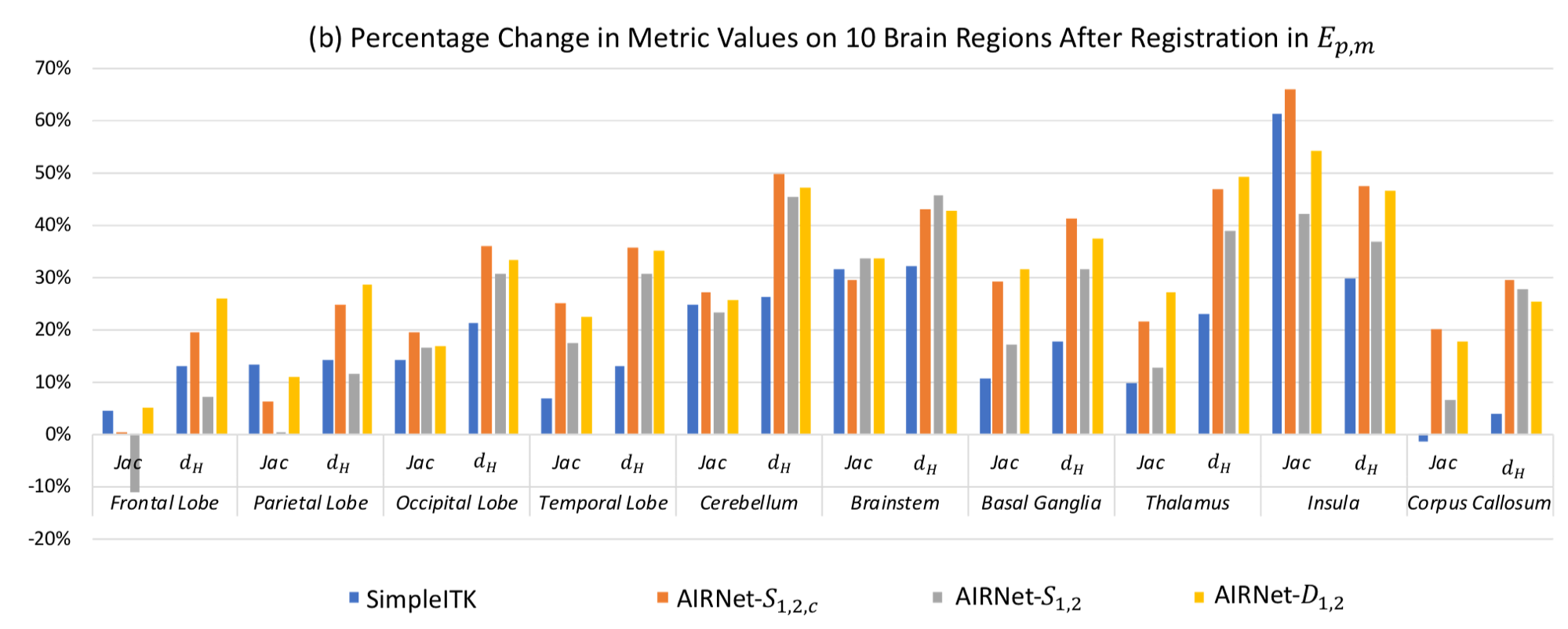} 
    \caption{Performance of the conventional method and our models when evaluated 
    on ten regions of the brain. The charts show the percentage change in the mean
    Jaccard index and modified Hausdorff distance on each region when compared to 
    before registration}
    \label{fig:ten_reg}
\end{figure}

Next, we evaluate the performance of the methods on ten out of the eighteen
brain regions which are labelled. We note that in some methods, only affine 
transformation is involved and hence, not all of the brain parts will be registered 
perfectly. The registration performance could vary across different brain regions and 
the results are shown in Fig.~\ref{fig:ten_reg} and Table~\ref{table:ten_reg_sum}. 

\setlength{\tabcolsep}{4pt}
\begin{table}
    \caption{Average performance of the conventional methods and our proposed 
    framework summarised from Fig.~\ref{fig:ten_reg}. [left] Average percentage 
    improvement in the metrics across ten brain parts when compared to before 
    registration. [right] Number of regions which our proposed method performs better 
    than the conventional method in terms of the respective metrics. 
    The best results are highlighted in bold.}
    \label{table:ten_reg_sum}
    \begin{minipage}{.55\linewidth}
      \centering
        \begin{tabular}{lcccc}
        \hline\hline
        \noalign{\smallskip}
        \multirow{2}{*}{Method} & \multicolumn{2}{c}{$E_p$} & \multicolumn{2}{c}{$E_{p,m}$} \\
        \noalign{\smallskip}
        \cline{2-5}
        \noalign{\smallskip}
        & $Jac$ & $d_H$ & $Jac$ & $d_H$ \\
        \noalign{\smallskip}
        \hline\hline
        \noalign{\smallskip}
        SimpleITK \cite{simpleitk,simpleitk2} & 25.5 & 23.8 & 17.6 & 19.6 \\
        AIRNet-$S_{1,2,c}$ & \bf{28.6} & \bf{39.8} & 24.5 & \bf{37.5} \\
        AIRNet-$S_{1,2}$ & 19.4 & 32.3 & 15.9 & 30.7 \\
        AIRNet-$D_{1,2}$ & 21.9 & 34.6 & \bf{24.6} & 37.2 \\
        \noalign{\smallskip}
        \hline\hline
        \end{tabular}
    \end{minipage}%
    \begin{minipage}{.45\linewidth}
      \centering
        \begin{tabular}{lcccc}
        \hline\hline
        \noalign{\smallskip}
        \multirow{2}{*}{Method} & \multicolumn{2}{c}{$E_p$} & \multicolumn{2}{c}{$E_{p,m}$} \\
        \noalign{\smallskip}
        \cline{2-5}
        \noalign{\smallskip}
        & $Jac$ & $d_H$ & $Jac$ & $d_H$ \\
        \noalign{\smallskip}
        \hline\hline
        \noalign{\smallskip}
        AIRNet-$S_{1,2,c}$ & \bf{5} & \bf{10} & 7 & \bf{10} \\
        AIRNet-$S_{1,2}$ & 4 & 7 & 6 & 8 \\
        AIRNet-$D_{1,2}$ & 3 & 9 & \bf{8} & \bf{10} \\
        \noalign{\smallskip}
        \hline\hline
        \end{tabular}
    \end{minipage} 
\end{table}
\setlength{\tabcolsep}{4pt}

On the whole, it is observed that except for AIRNet-$D_{1,2}$, all other 
methods perform better in across patients registration, $E_p$. In fact, the 
conventional algorithm displays the most significant improvement in terms of 
the metric values, suggesting that it is more favourable for registration tasks on 
images of the same modality. As opposed to the other models, AIRNet-$D_{1,2}$ 
was trained on image pairs of different modalities, and hence is reasonable to 
perform better in $E_{p,m}$. However, when compared with AIRNet-$S_{1,2,c}$, 
there was only a slight advantage in performance. Our framework using 
AIRNet-$S_{1,2,c}$ is still the overall best and has achieved better results in 
majority of the ten brain regions. This self-supervised learning method would be 
preferable because during the dataset preparation, it only requires one image for 
each training sample instead of two aligned images of different modalities. In 
addition, it is more flexible in terms of the number of modalities that can be included. 
Lastly, we observe that even though T1-weighted MR images were not included 
in the training of AIRNet-$S_{1,2}$, it still manages to perform reasonably well on 
the test cases which involve this image modality, indicating that the network is 
able to recover the same features from the unseen data type. 

To summarise, our proposed framework based on self-supervised learning can 
register images significantly faster than the conventional method and can achieve 
superior performance both in terms of $Jac$ and $d_H$. Although trained on image 
pairs of the same patient and modality, the model still works well on images without 
such properties. Thus, our method can be applied to registration of images from 
different patients and/or modalities. The AIRNet also provides decent registration 
results when tested on images of unseen modalities, indicating that the network 
encoder is extracting generalised features of the image regardless of its modality. 
We visualise and further verify this inference in the next part.  

\subsection{Encoder Output Visualisation}
 
In this part, we examined the output of the encoder of our best performing model
AIRNet-$S_{1,2,c}$. For each input image, we extract the output of the flattening 
layer in the network, which would be the new representation of this image. To 
visualise this high dimensional output, we use principal component analysis (PCA) 
\cite{pca1,pca2} and t-Distributed Stochastic Neighbour Embedding (t-SNE) \cite{tsne}.

\begin{figure}
    \centering
    \includegraphics[width=0.65\textwidth]{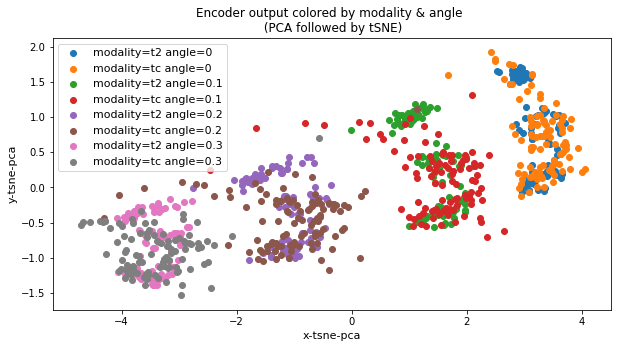} 
    \includegraphics[width=0.65\textwidth]{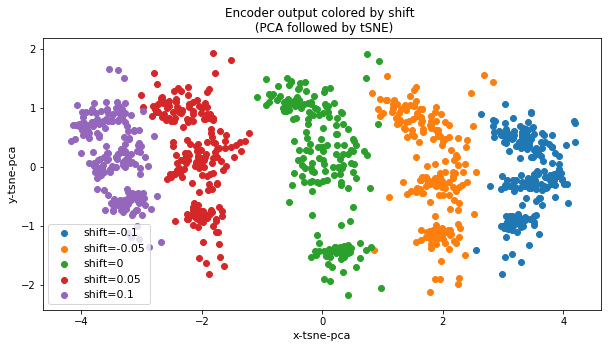} 
    \caption{Scatter plot of encoder outputs after dimensionality-reduction using
    PCA, followed by tSNE. [above] Each sample is labeled by the angle (in radians) 
    that it is rotated and the type of modality ({\it t2} is T2-weighted and {\it tc} is 
    contrast-enhanced T1-weighted). [below] Each sample is categorised by the 
    type of translation used}
    \label{fig:tsne}
\end{figure}

First, the representation of MR images with the brains being rotated at various 
angles were extracted and visualised in the first image of Fig.~\ref{fig:tsne}. 
Samples are clustered according to their rotation angle, indicating that the 
encoder is able to incorporate information regarding the angle of rotation for each 
image. Although T2 and contrast-enhanced T1-weighted images are included, 
there is no significant boundaries between modalities, emphasising that the 
encoded representation are classified mainly based on the transformation 
instead of modality. This suggests that the encoder views the same brain scan 
of different modalities similarly and focuses on extracting geometric information. 
Similarly, the second image in Fig.~\ref{fig:tsne} visualises the clusters of the 
encoder output based on the amount of shift applied onto the images. The 
samples of different modalities are not labelled explicitly but there is also no 
obvious boundary between them as observed previously.

 \begin{figure}
    \centering
    \includegraphics[width=0.65\textwidth]{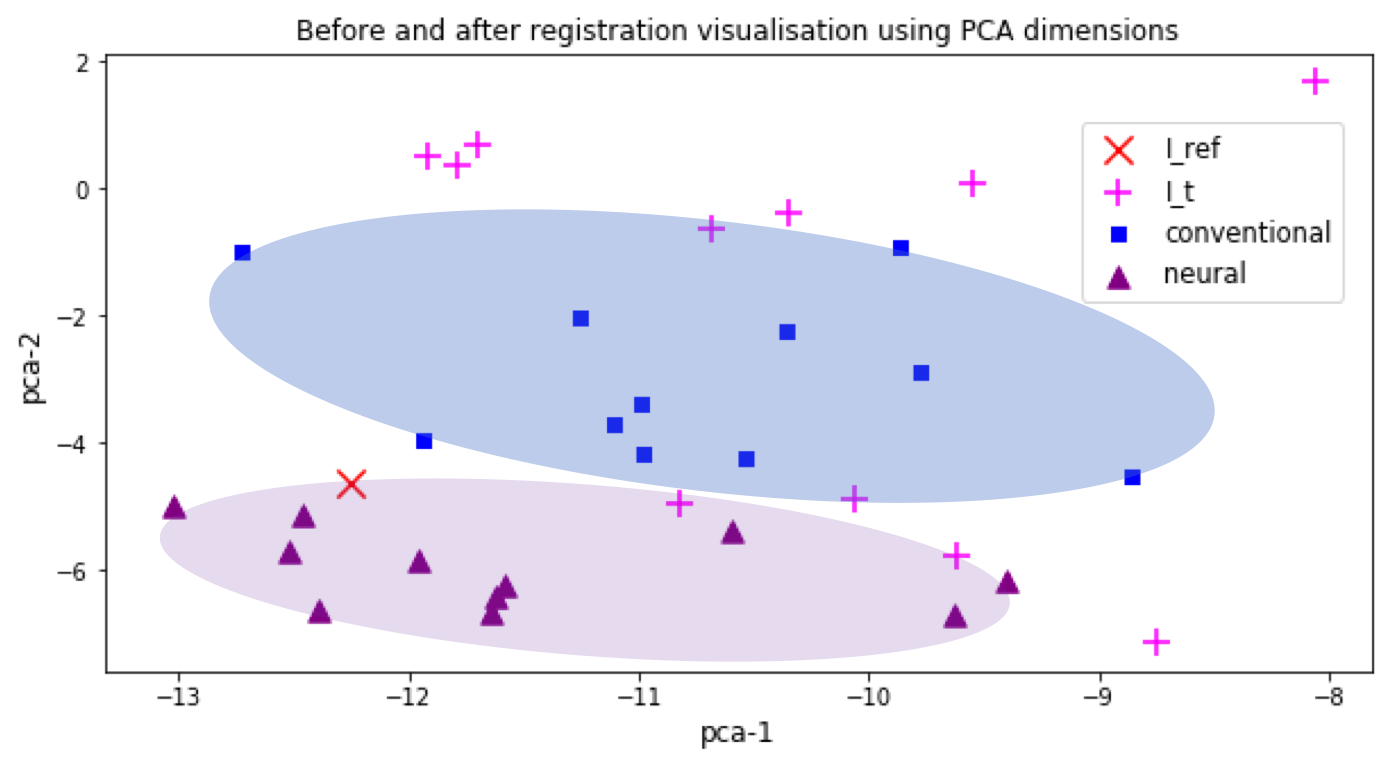} 
    \caption{Scatter plot of encoder outputs of images from $E_{p,m}$ after 
    dimensionality-reduction using PCA. The red and pink markers represent the 
    reference image $I_{ref}$ and images to be registered $I_{\vec{t}}$ respectively. 
    The registration results using AIRNet-$S_{1,2,c}$ and SimpleITK are indicated 
    by the purple and blue markers}
    \label{fig:pca_perf_cluster}
\end{figure}

Additionally, we visualise the trajectory of the registration results of our proposed 
method and the conventional algorithm. Fig.~\ref{fig:pca_perf_cluster} shows an 
example of such visualisation illustrating the images before and after registration 
on the eleven subjects from $E_{p,m}$. All images were passed through the encoder 
to obtain their representation. For both methods, the registered images are more 
clustered and have moved towards the reference image in terms of PCA dimensions. 
Moreover, it is observed that the results from our proposed framework are closer to 
the target, agreeing with the performance shown in Table~\ref{table:ten_reg_sum}. 

Taken together, the results suggest that the encoder is useful in selecting 
discriminative features that describe complex patterns in the images and provide 
important information to perform the image registration task. 

\section{Conclusions}

A self-supervised learning method for affine 3D image registration has been 
presented and evaluated on medical images, specifically on axial view of brain 
scans. The presented system achieves a superior performance compared to a 
conventional image registration method at a much shorter execution time. 

Since it is time-consuming and tedious to obtain information regarding the 
transformation between every two 3D images, we created a synthetic dataset 
using the existing medical images so that for each images, we have a new 
transformed image and a label for the pair without the need of manual annotation. 
With these data, we proceed with the training of the AIRNet in a self-supervised 
manner. Although each input pair of the training data are of the same patient and
same modality, the trained network were tested for registration across patients 
and modalities and it still manages to perform well as compared with the 
conventional method.

Besides having a network which is able to find an affine transformation between
two images, the way the AIRNet is constructed enables us to find a good
feature representation of the images. This representation provides a concise
description of the complex morphological patterns of the images regardless of 
the modality of the input images. 

To sum up, our experiments show good registration results for images between
different patients and different modalities by extracting useful features of the 
input images. To extend the applicability of the proposed method in future work, 
performing deformable registration will be investigated. Moreover, experiments 
were performed using only axial view of brain scans, but this can be extended 
to register any other type of 3D images.

\bibliographystyle{is-unsrt}
\bibliography{airnet_ref}

\end{document}